\documentclass{article}

\usepackage{blindtext} 

\usepackage[sc]{mathpazo} 
\usepackage[T1]{fontenc} 
\usepackage{microtype} 

\usepackage[english]{babel} 

\usepackage[hmarginratio=1:1,top=32mm,columnsep=20pt]{geometry} 
\usepackage[hang, small,labelfont=bf,up,textfont=it,up]{caption} 
\usepackage{booktabs} 

\usepackage{lettrine} 

\usepackage{enumitem} 
\setlist[itemize]{noitemsep} 

\usepackage{abstract} 

\usepackage{titlesec} 
\renewcommand\thesection{\Roman{section}} 
\renewcommand\thesubsection{\roman{subsection}} 
\titleformat{\section}[block]{\large\scshape\centering}{\thesection.}{1em}{} 
\titleformat{\subsection}[block]{\large}{\thesubsection.}{1em}{} 

\usepackage{fancyhdr} 
\usepackage{titling} 

\usepackage{hyperref} 

\newcommand{\argmin}{\mathop{\rm arg~min}\limits}

\usepackage{cite}

\pretitle{\begin{center}\Huge\bfseries} 
\posttitle{\end{center}} 
\title{Tuning Support Vector Machines by Iterated Local Search} 
\author{%
\textsc{Sergio Consoli}\thanks{Corresponding author} \\[1ex] 
\normalsize Philips Research \\ 
\normalsize \href{mailto:sergio.consoli@philips.com}{sergio.consoli@philips.com} 
\and 
\textsc{Jacek Kustra} \\[1ex] 
\normalsize Philips Research \\ 
\normalsize \href{mailto:jacek.kustra@philips.com}{jacek.kustra@philips.com} 
\and 
\textsc{Pieter Vos} \\[1ex] 
\normalsize Philips Research \\ 
\normalsize \href{mailto:pieter.vos@philips.com}{pieter.vos@philips.com} 
\and 
\textsc{Monique Hendriks} \\[1ex] 
\normalsize Philips Research \\ 
\normalsize \href{mailto:monique.hendriks@philips.com}{monique.hendriks@philips.com} 
\and 
\textsc{Dimitrios Mavroeidis} \\[1ex] 
\normalsize Philips Research \\ 
\normalsize \href{mailto:dimitrios.mavroeidis@philips.com}{dimitrios.mavroeidis@philips.com} 
}
\date{\today} 
\renewcommand{\maketitlehookd}{%
\begin{abstract}
We provide preliminary details and formulation of an optimization strategy under current development that is able
to automatically tune the parameters of a Support Vector Machine over new datasets.
The optimization strategy is a heuristic based on Iterated Local Search, a modification of classic hill climbing which iterates calls to a local search routine. 
\end{abstract}
}

\begin{document}

\maketitle


\section{Introduction} \label{introduction}

The performance of Support Vector Machine (SVM) strongly relies on the initial setting of the model parameters \cite{Vapnik2000}.
The parameters are usually set by training the SVM on a specific dataset and are then fixed when applied to a certain application.
The automatic configuration for algorithms 
is faced with the same problem when doing hyper-parameter tuning in machine learning: finding the optimal setting of those parameters is an art by itself and as such much research on the topic has been explored in the literature~\cite{ceyl16,lame15,yang12,sher15,smac2011}.
Of the techniques used, grid search (or parameter sweep) is one of the most common methods to approximate optimal parameter values~\cite{gridsearch}.
Grid search involves an exhaustive search through a manually specified subset of the hyperparameter space of a learning algorithm, guided by some performance metric (e.g. cross-validation).
This traditional approach, however, has several limitations: (i) it is vulnerable to local optimum; (ii) it does not provide any global optimality guarantee;
(iii) setting an appropriate search interval is an ad-hoc approach; (iv) it is a computationally expensive approach, especially when search intervals require to capture wide ranges.

In this short contribution, we describe our preliminary investigation into an optimization method which tackles the parameter setting problem in SVMs using Iterated Local Search (ILS) \cite{Lourenco2010},
a popular explorative local search method used for solving discrete optimization problems.
It belongs to the class of trajectory optimization methods, i.e. at each iteration of the algorithm the search process designs a trajectory in the search space, starting from an initial state and dynamically
adding a new, better solution to the curve in each discrete time step.
Iterated Local Search mainly consists of two steps. In the first step, a local optimum is reached by performing a walk in the search space, referred to as the \textit{perturbation phase}.
The second step is to efficiently escape from local optima by using an appropriate \textit{local search phase} \cite{lour1995}. 
The application of an \textit{acceptance criterion} to decide which of two local candidate solutions has to be chosen to continue the search process is also an important aspect of the algorithm. 
%
In the next section we provide the preliminary details and formulation of our optimization strategy based on ILS for the parameter tuning task in SVMs.

\section{ILS for SVM parameters tuning} \label{method}

Given the input parameters $x\in X$ and their corresponding output parameters $y \in Y = \{-1,1\}$,
the separation between classes in SVMs is achieved by fitting the hyperplane $f(x)$ that has the optimal distance to the nearest data point used for training of any class: $f(x) = \sum_{i=1}^{n}{\alpha_i y_i <x_i,x>+b}$, where $n$ is the total number of parameters. The goal in SVMs is to find the hyperplane which maximizes the minimum distances of the samples on each side of the plane~\cite{svma1}. 
A penalty is associated with the instances which are misclassified and added to the minimization function. This is done via the parameter $C$ in the minimization formula: $\argmin\limits_{f(x)=\omega^T x+b}\frac{1}{2}\|\omega\|^2+C\sum_{i}^{n}c(f, x_i, y_i)$.

By varying $C$, a trade-off between the accuracy and and stability of the function is defined. Larger values of $C$ result in a smaller margin, leading to potentially more accurate classifications, however overfitting can occur.
A mapping of the data with appropriate kernel functions $k(x, x')$ into a richer feature space, including non-linear features is applied prior to the hyperplane fitting. Among several kernels in the literature, 
we consider the Gaussian radial-basis function (RBF): $K(x_i,x')=exp(-\gamma\|x_i-x'\|^2), \gamma>0$, where $\gamma$ defines the variance of the RBF, practically defining the shape of the kernel function peaks: lower $\gamma$ values set the bias to low and corresponding high $\gamma$ to high bias.

The proposed ILS under current implementation for SVM tuning uses grid search \cite{gridsearch} as an inner local search routine, which is then iterated in order to make it fine-grained 
and finally producing the best parameters $C$ and $\gamma$ found to date. 
Given a training dataset $D$ and an SVM model $\Theta$, the procedure first generates an initial solution. 
We use an initial solution produced by grid search. The grid search exhaustively generates candidates from a grid of the parameter values, $C$ and $\gamma$, specified in the arrays $range_{\gamma} \in \Re ^+$ and $range_C \in \Re ^+$. We choose arrays containing five different values for each parameter, so that the grid search method will look to $25$ different parameters combinations. The range values are taken as different powers of 10 from $-2$ to $2$. 
Solution quality is evaluated as the accuracy of the SVM by means of $k$-fold cross validation \cite{McLachlan2004}, and 
stored in the variable $Acc$.

Afterwards, the \textit{perturbation phase}, which
represents the core idea of ILS, is applied to the incumbent solution. The goal is to provide a good starting point (i.e. parameter ranges) for the next \textit{local search phase} of ILS (i.e. the grid search in our case), based on the previous search experience of the algorithm, so as to obtain a better balance between exploration of the search space against wasting time in areas that are not giving good results. Ranges are set as: $ range_{\gamma} = [\gamma * 10^{-2}, \gamma * 10^{-1}, \gamma, \gamma * 10, \gamma * 10^{2}] \equiv [\gamma_{inf-down}, \gamma_{inf-up}, \gamma, \gamma_{sup-down}, \gamma_{sup-up}]$, and $range_{C} = [C * 10^{-2}, C * 10^{-1}, C, C * 10, C * 10^{2}] \equiv [C_{inf-down}, C_{inf-up}, C, C_{sup-down}, C_{sup-up}]$.
\\
\\
Imagine that the grid search gets the set of parameters $\gamma', C'$ as a new incumbent solution, whose evaluated accuracy is $Acc'$. 
Then the \textit{acceptance criterion} of this new solution is that it produces a better quality, that is an increased accuracy, than the best solution to date. If it does not happen, the new incumbent solution is rejected and the ranges are updated automatically with the following values: $\gamma_{inf-down}= rand(\gamma_{inf-down} * 10^{-1}, \gamma_{inf-down}) \textrm{ and } C_{inf-down}= rand(C_{inf-down} * 10^{-1}, C_{inf-down})$,  $\gamma_{inf-up}= rand( \frac{\gamma - \gamma_{inf-up}}{2}, \gamma) \textrm{ and } C_{inf-up}= rand( \frac{C - C_{inf-up}}{2}, C)$, $\gamma_{sup-down}= rand( \frac{\gamma_{sup-down} - \gamma}{2}, \gamma) \textrm{ and } C_{sup-down}= rand( \frac{C_{sup-down} - C}{2}, C)$, and $\gamma_{sup-up}= rand(\gamma_{sup-up} * 10) \textrm{ and } C_{sup-up}= rand(C_{sup-up} * 10)$.
That is, indifferently for $\gamma$ and $C$, the values of the $inf$-$down$ and $sup$-$up$ components are random values always taken farther the current parameter ($\gamma$ or $C$), in order to increase the diversification capability of the metaheuristic; while the values of the $inf$-$up$ and $sup$-$down$ components are random values always taken closer the current parameter, in order to increase the intensification strength around the current parameter. This perturbation setting should allow a good balance among the intensification and diversification factors.

Otherwise, if in the acceptance criterion the new incumbent solution, $\gamma'$ and $C'$, is better than the current one, $\gamma$ and $C$, i.e. $Acc' > Acc$, then this new solution becomes the best solution to date ($\gamma \leftarrow \gamma'$, $C \leftarrow C'$), and $range_{\gamma}$ and $range_C$ are updated as usual. 
This procedure continues iteratively until the termination conditions imposed by the user 
are satisfied, producing at the end the best combination of $\gamma$ and $C$ as output.



\section{Summary and outlook} \label{conclusions}

We considered the parameter setting task in SVMs by an automated ILS heuristic, which looks to be a promising approach. 
We are aware that a more detailed description of the algorithm is deemed necessary, along with a thorough computational investigation. This is currently object of ongoing research,
including a statistical analysis and comparison of the proposed algorithm against the standard grid search, in order to quantify and qualify the improvements obtained. Further
research will explore the application of this strategy to other SVM kernels, considering also a variety of big, heterogenous datasets.

\begin{footnotesize}
\bibliographystyle{splncs}
\bibliography{bibliofile}
\end{footnotesize}


\end{document}